\def\year{2018}\relax
\documentclass[letterpaper]{article} %DO NOT CHANGE THIS
\usepackage{aaai18}  %Required
\usepackage{times}  %Required
\usepackage{helvet}  %Required
\usepackage{courier}  %Required
\usepackage{url}  %Required
\usepackage{graphicx}  %Required
\usepackage{amssymb}
\usepackage{enumerate}
\usepackage{mathtools}
\usepackage{amsmath}
\usepackage{color}
\usepackage{blindtext}
\usepackage{url}
\usepackage{csquotes}
\usepackage{algorithm,algcompatible}
\algnewcommand\INPUT{\item[\textbf{Input:}]}%
\algnewcommand\OUTPUT{\item[\textbf{Output:}]}%
\usepackage{tabularx}
\usepackage{subcaption}
\usepackage{flushend} % This makes the columns on the last page equal in length.
\usepackage{multirow} %For multirow in tables
\usepackage[colorinlistoftodos]{todonotes}
\begin{document}

\title{Apprenticeship Bootstrapping via Deep Learning with a Safety Net for UAV-UGV Interaction}
\author{H. Nguyen, V. Tran, T. Nguyen, M. Garratt, K. Kasmarik, M. Barlow, S. Anavatti, and H. Abbass\\
School of Engineering \& IT, UNSW Canberra, Australia
}

\maketitle

\begin{abstract}

In apprenticeship learning (AL), agents learn by watching or acquiring human demonstrations on some tasks of interest. 
However, the lack of human demonstrations in novel tasks where they may not be a human expert yet, or when it is too expensive and/or time consuming to acquire human demonstrations motivated a new algorithm: Apprenticeship bootstrapping (ABS). The basic idea is to learn from demonstrations on sub-tasks then autonomously bootstrap a model on the main, more complex, task. The original ABS used inverse reinforcement learning (ABS-IRL). However, the approach is not suitable for continuous action spaces.

In this paper, we propose ABS via Deep learning (ABS-DL). It is first validated in a simulation environment on an aerial and ground coordination scenario, where an Unmanned  Aerial Vehicle (UAV) is required to maintain three Unmanned  Ground Vehicles (UGVs) within a field of view of the UAV 's camera (FoV). Moving a machine learning algorithm from a simulation environment to an actual physical platform is challenging because `mistakes' made by the algorithm while learning could lead to the damage of the platform. We then take this extra step to test the algorithm  in a physical environment. We propose a safety-net as a protection layer to ensure that the autonomy of the algorithm in learning does not compromise the safety of the platform. The tests of ABS-DL in the real environment can guarantee a damage-free, collision avoidance behaviour of autonomous bodies. The results show that performance of the proposed approach is comparable to that of a human, and competitive to the traditional approach using expert demonstrations performed on the composite task. The proposed safety-net approach demonstrates its advantages when it enables the UAV to operate more safely under the control of the ABS-DL algorithm.

\end{abstract} 

\section{Introduction} \label{sec:1}

Designing a reward function for a reinforcement learning agent could be a cumbersome task. Using human experts to demonstrate a task to an artificial agent to learn from could both speed up the learning process and equally reduces the burden of designing reward functions by hand. However, even this solution is not as simple as it may sound.

In recent surveys \cite{argall2009survey}, \cite{billing2010formalism}, \cite{hussein2017imitation}, the main challenges emanate from the problem of how to transfer human skills to  agents or robots through demonstrations. When designing a new task for an autonomous system, particularly in complex situations or tasks, there is no guarantee that there is a human expert or, if so, that he/she is available to create a dataset for the robot. 

The previous challenge called for designing a new learning scheme, called Apprenticeship Bootstrapping (ABS) for learning a composite task using human demonstrations of sub-tasks~\cite{hung2018ABS,hung2018ABSIJCNN}. An ABS via inverse reinforcement learning algorithm (ABS-IRL) has shown success in overcoming that challenge. However, it is not suitable for continuous action spaces. This motivated us to propose a new ABS approach via deep learning, called ABS-DL, which is described in the next section.

The validation task is designed to mimic the simulated task in previous work on ABS, which was a ground-air interaction scenario~\cite{hung2017supervised,hung2018ABS,hung2018ABSIJCNN}. The aerial and ground coordination task is a challenge in order to control the UAV for the human operator. Therefore, the task is suitable to evaluate our ABS-DL when it is decomposed into sub-tasks and then the proposed ABS-DL algorithm is used to learn from these sub-tasks before application to a composite task. 

However, when applying our ABS-DL algorithm for physical environments, it is challenging to overcome the safety concerns especially when there is no human involved in the operation. Therefore, in this paper, we propose a primary safety-net approach to limit the UAV behaviour produced by our ABS-DL algorithm. 

We first present previous work on safety nets. This is then followed by a description of the proposed ABS-DL algorithm. The scenario used for evaluating the algorithm is then presented, followed by experiments in both the simulated and physical environments and results. Conclusions are then drawn.

\section{Safety Nets for Learning Agents} \label{sec:3}

When we apply learning algorithms in real-world operations, we cannot afford to overlook the safety issues such as damages to humans and systems in the environment caused by the errors of the algorithm. Most systems in the academic literature for autonomous systems, unmanned  vehicles, and human-robot interactions rely entirely on the output generated by optimal control or machine learning algorithms without safety nets, which limits the applicability of those methods in practice \cite{chaulwar2017machine,geng2018study,zhan2017spatially}. Recently, many researchers started to include constraints that are used to limit the action or behaviour produced by autonomy within a safe zone, so that the tests of the models in the real environments can guarantee a damage-free collision avoidance behaviour. These terms are frequently called \emph{safety nets}, \emph{safety margins} or \emph{safety constraints}. While there are subtle differences in the meaning, we prefer Safety nets as they represent the overall architecture and system that contains the safety constraints and safety margins.

There are three types of safety nets that have been used in recent studies to limit the action or to guide the learning, including internal hard constraints, internal soft constraints, and external intervention protocols.

\begin{itemize}
\item \textbf{Internal hard constraints} are the constraints in the methodology besides the algorithms themselves providing one or more safety margins that use rules to limit the output actions. These constraints are found in very simple rule-based form in many path planning and UAV applications. For instance, the hard constraints can be defined as the minimum distance to obstacles, and maximum velocity of the vehicles \cite{chae2017probabilistic,chen2017double,miraglia2017dynamic}. In \cite{raineri2017online}, small and large safety margins have been introduced to add extra layers of safety before the planning algorithms produce collision-free trajectories. In some circumstances, the internal hard constraints can be found in a form of predefined parameters which are later used to compute the dynamic safety margins \cite{mayer2017planning,suh2018stochastic}.

\item \textbf{Internal soft constraints} take forms of additional learning algorithms to compute the dynamic safety margins. Instead of defining a fixed constrained distance or kinematic-related parameters to deduce safety margins for collision avoidance, extra predictive models might be used to fuse the environmental and user-related information to compute dynamic safety constraints that regularize the output of the algorithms \cite{arbabzadeh2018data,hubschneider2017integrating}.

\item \textbf{External intervention protocols} are the ultimate safety nets that can be used by humans to intervene manually to avoid dangerous situations. The autonomous systems can be equipped with an active safety mode that can grant the human operators the overriding right to make appropriate decisions given the lack of autonomy to handle the situations \cite{khan2017autonomous,Punzo2018}.
\end{itemize}

Our experiments are performed in simulation as well as in physical environments, which consists of an UAV and multiple UGVs. To avoid any damage to the UAV and UGVs in the physical environment, which is not a significant issue when training in simulation, we adapt internal hard constraints and external intervention protocols as safety nets for our tests. Details of those safety nets are discussed below.

\subsection{Safety nets for the UAV and the UGV}\label{sec:safetynets}

In this paper, to achieve safe experimental conditions, we introduce a double-layer safety net including motion hard constraints and external intervention protocols. Both of the constraints and protocols are used in our physical experiment. The safety net demonstrates its benefits when it helps the testing experiment of our ABS-DL algorithm on the physical environment to avoid damage for the UAV, the UGVs, and the surrounding environment and obstacles.

Let $o_j\in \mathcal{O}$ be the obstacle $j$ in a set of fixed obstacles $\mathcal{O}$. Internal hard constraints are enforced on the future positions of the UAV and UGVs according to the following equation:

\begin{equation}
\eta = \begin{cases} 1, & \mbox{if } x + \xi > x_{o_j}^- \\ & \mbox{if } x - \xi < x_{o_j}^+ \\ & \mbox{if } y + \xi > y_{o_j}^- \\ & \mbox{if } y - \xi < y_{o_j}^+ \\ 0, & \mbox{otherwise} \end{cases} \qquad \forall o_j\in \mathcal{O}
\end{equation}

where \( x_{o_j}^-,x_{o_j}^+ \) denote the x-coordinates of obstacle $j$' left edge and right edge, respectively; \( y_{o_j}^-,y_{o_j}^+ \) denote the y-coordinates of obstacle $j$' upper edge and lower edge. $\eta = 1$ indicates the unsafe condition of the  UAV or UGV position relative to obstacles, which activates the hover action of UAV or the immobilization of the UGV. $\xi$ determines the thickness of the safety margins.

There is also a constraint with thickness of $\xi$ at four boundaries of the testing area. The distance between the UAV or any UGV and each boundary of the environment or any edge of an obstacle is estimated based on sensor data. If the safety margin is predicted to be crossed over by the next action of any vehicle, the violated vehicle will be forced to hover at one point, in the case of UAV, or to stop moving, in the case of UGV, to guarantee a collision-free trajectory (collision with the environment boundary). The vehicles then wait for the next non-violation action or the command of the human operator.

Another layer of the safety net is the manual mode that can be selected by the human operator to override and control the UAV and UGV in hazardous situations when the human notices any issue in taking off, landing, or any risk of collisions.

\section{Apprenticeship Bootstrapping via Deep Learning for Robotics} \label{sec:2:2}

A motion tracking system generally produces the desired movement via the robot's motion capability on the horizontal and vertical axes by time. Thanks to this technique, the corresponding robot is driven to follow its own predefined trajectory until it reaches the target point. Then, the formation pattern between robots is generated and maintained during the transient process \cite{wenzel2011automatic,yu2015cooperative}. However, the development of a robot motion planner is not an easy task since this problem involves many complicated and mixed steps. Firstly, system model identification for active UAV and UGV systems must be considered based on observing input-output signals from experimental data. This dynamic analysis of given systems provides controller designers with a better understanding of the system behaviour, notably the cause-effect relationships \cite{koszewnik2014parrot,bouabdallah2007full,phan2008cooperative}. After a proper model is determined, the linear and angular velocity tracking control diagrams on the longitudinal and vertical axes are designed and implemented in each robot with the aim of stabilizing the robot's velocity by PID controllers as much as possible throughout its movement. 

In this paper, to reduce the complexity of planning motions between UAV and UGVs, we develop an artificial intelligence (AI) controller for the UAV. Additionally, we integrate our proposed safety nets for the control of the UAV agent in this task. This integration allows the UAV avoid unexpected behaviours produced by the agent.

One approach to design the AI controller is to use human experts. However, in practice, a human expert may not be available because the tasks are new or it is expensive to access someone with the required skills to perform the task. By decomposing the skill into sub-skills that require less skilled humans, we can bootstrap the higher skills from these building blocks. This has been the primary motivation for ABS-DL. 

The sub-skills represent a decomposition of the action space. Not all actions are needed for a sub-skill. It may also involve a decomposition of the state space since sub-skills are associated with simpler contexts that represent partial representations of the original context. Below, we will explain the above formally.

Define $S= \{ S_1,S_2,...,S_N \}$ and $A= \{ A_1, A_2, ..., A_L \}$ to be the original complex state and action spaces of a complex task, respectively. Here, $S_n$ and $A_l$ represent the sub-state and sub-action spaces. Supposing that the composite task is divided into $H$ sub-tasks.

Let $D= \{D_1, D_2,..., D_H\}$ be a set of demonstrations of all sub-tasks, where $D_h$ is a set of $h$ sub-task demonstrations. Each set of $h$ sub-task demonstrations, $D_h$, is comprised of state-action pairs $(s_t,a_t)$. To form the set $D_h$, in the sub-state space $S_h$, the expert is required to perform actions in the sub-action space $A_h$; meanwhile $S_h\subseteq S$ is one of the sub-state spaces $S_n$ in the whole complex state space, and $A_h\subseteq A$ is one of sub-action spaces $A_l$ in the whole complex action space. It is important to emphasize that the sub-tasks are orthogonal. 

In this learning approach, it is assumed that the number of dimensions of the states in all sub-state spaces is identical, and the composite action space is decomposed into different sub-action spaces. While performing the sub-task, the expert is required to focus on only the dimensions relevant to this sub-task, and use the corresponding sub-action space. 

To create a composite set from the sub-task demonstrations, a straight-forward fusion of all sub-state spaces that have the identical number of dimensions, and possibly primitive actions of sub-action spaces are combined to produce a sufficient action space similar to the composite action space of the complex task.

A deep network is used to train the composite set. States of sub-state spaces are input for the network, and the sufficient action space covering all possible primitive actions is the network response. The high-level algorithmic description of ABS-DL is shown in Algorithm~\ref{algh-ABS-DL}.

\begin{algorithm}[h!]
  \caption{Apprenticeship Bootstrapping (ABS) via Deep Learning Algorithm.}\label{algh-ABS-DL}
  \begin{algorithmic}[1]   
     \INPUT Sub-task demonstrations $\{D_1, D_2, D_3, ..., D_H\} $; $F_{state}$ - the function fusing a sub-state into a composite state; $F_{action}$ - the function fusing a sub-action into a composite action
    \OUTPUT A DNN trained model outputting composite actions.
        \STATE Initializing a composite set.
        \STATE Initializing a DNN model.
        \FOR{each sub-task demonstration ($s,a$)} 
            \STATE $s_{composite} = F_{state}(s)$
            \STATE $a_{composite} = F_{action}(a)$
            \STATE Adding the composite demonstration  ($s_{composite}, a_{composite}$) to the composite set.
         \ENDFOR
        \STATE Training the DNN model using the composite set. 
  \end{algorithmic}
\end{algorithm}

\section{EXPERIMENTAL APPARATUS}\label{sec:5}
\subsection{UAV/UGVs Coordination Task}
\begin{figure}[h]
\centering
\includegraphics[width=1.0\linewidth]{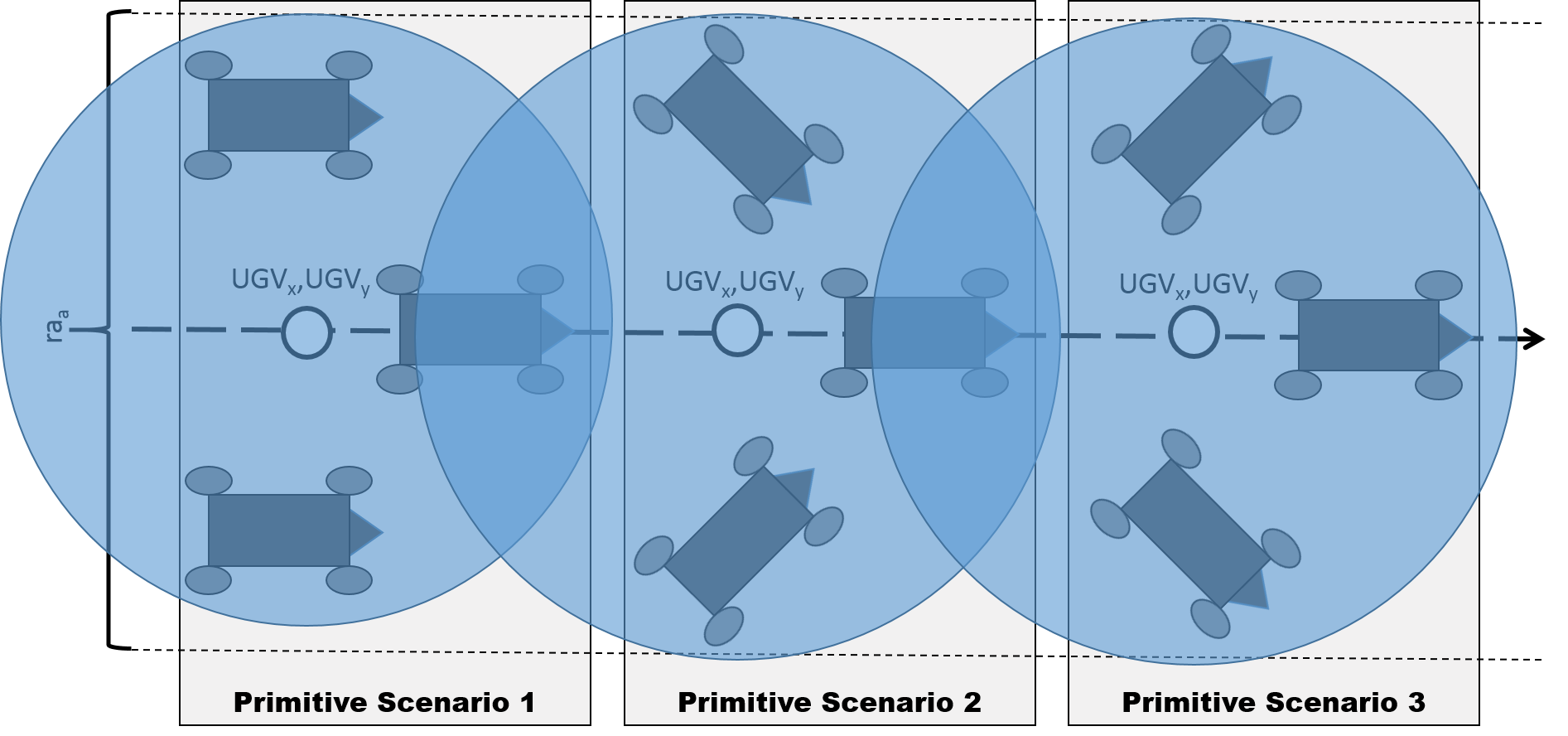}
\caption{Formation of UGVs  in three primitive scenarios.}
\label{fig:UGVs}
\end{figure}

In this paper, the coordination task of an aerial-ground is identical to that in our previous research \cite{hung2017supervised,hung2018ABSIJCNN,hung2018ABS}. There are four manoeuvres: Fixed-Altitude manoeuvre, Climb manoeuvre, Descend manoeuvre, and Combined manoeuvre. The first three are primitive manoeuvres and the fourth a composite and more complex one requiring switching among the three primitive ones. The six basic actions for controlling the UAV are: roll (move left and move right), pitch (move forward and move backward), and yaw (climb and descend). The formations of the UGVs in each of these three primitive manoeuvres are shown in Figure~\ref{fig:UGVs}. In addition to the UGVs formation control approach, an obstacle avoidance algorithm is also applied to UGVs to avoid unexpected obstacles and to see the UAV's behavior response while the UGV's predefined formation is shurnk or extended. This paper focuses on designing the autonomous control of the UAV; thus, the obstacle avoidance algorithm for UGVs will be discussed deeply in another paper.

The UAV is required to maintain three UGVs within its field of view (FoV) without missing any or creating a much larger FoV than needed to accommodate the manifold created by the UGVs.  Then, the UAV's task is decomposed into two objectives. Firstly, it needs to minimise the distance between its own centre of mass and that of the UGVs within its FoV. Secondly, it has to minimise the difference between the radius of its camera's FoV and the ideal one required which is defined as the radius of the smallest circle to encapsulate the manifold formed by the UGVs. The Pinhole camera model \cite{sturm2014pinhole} is used to determine the central points and radius values.

Let $(x^{\mathcal{A}}_t,y^{\mathcal{A}}_t)$ and $(x^{\mathcal{G}}_t,y^{\mathcal{G}}_t)$ be the centre of masses of the UAV and the UGVs within the UAV's FoV, respectively, and $\widehat{\nu}_t$ and $\nu_t$ the radius of the UAV camera's FoV and the ideal one at time step $t$. The first objective is to minimise the distance error given by Equation~\ref{eq:objective1}, where $||.||_2$ denotes the $L^2$ norm, and the second objective is to minimise the radius error expressed by Equation~\ref{eq:objective2}.  

\begin{equation}\label{eq:objective1}
\delta d = \int_t ||(x^{\mathcal{A}}_t,y^{\mathcal{A}}_t) - (x^{\mathcal{G}}_t,y^{\mathcal{G}}_t)||_2
\end{equation}

\begin{equation}\label{eq:objective2}
\delta \nu = \int_t (\nu_t - \widehat{\nu}_t)^2
\end{equation}

\begin{figure}[h]
\centering
\includegraphics[width=1.0\linewidth]{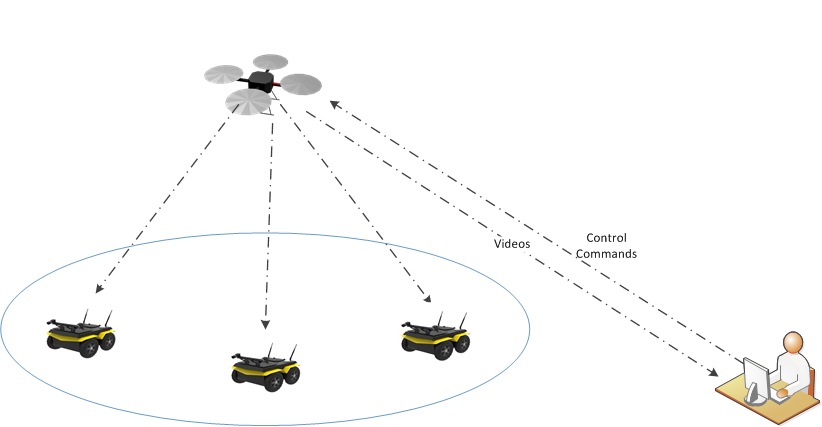}
\caption{Model of coordination task for UAV and UGVs}
\label{fig:environment}
\end{figure}

A pictorial representation of the overall architecture, where there are a human operator, an UAV, and there UGVs, is shown in Figure~\ref{fig:environment}. 

\subsection{Simulation Environment}
The Gazebo simulator~\cite{koenig2006gazebo} is used to design the task scenario and the drone simulator package in the Tum-Simulator~\cite{huang2014tum} is simulated for the Parrot AR. Drone 2.0. 

The simulation system allows a human to operate the UAV using a joystick. The operator aims to keep all UGVs in the field of view of the UAV's downward looking camera and centred in the image by watching the video telemetry continuously and making corrections with the joystick accordingly. Human demonstrations are collected and used as a data set for training our ABS-DL algorithm.

\begin{figure}[h]
\centering
\includegraphics[width=1.0\linewidth]{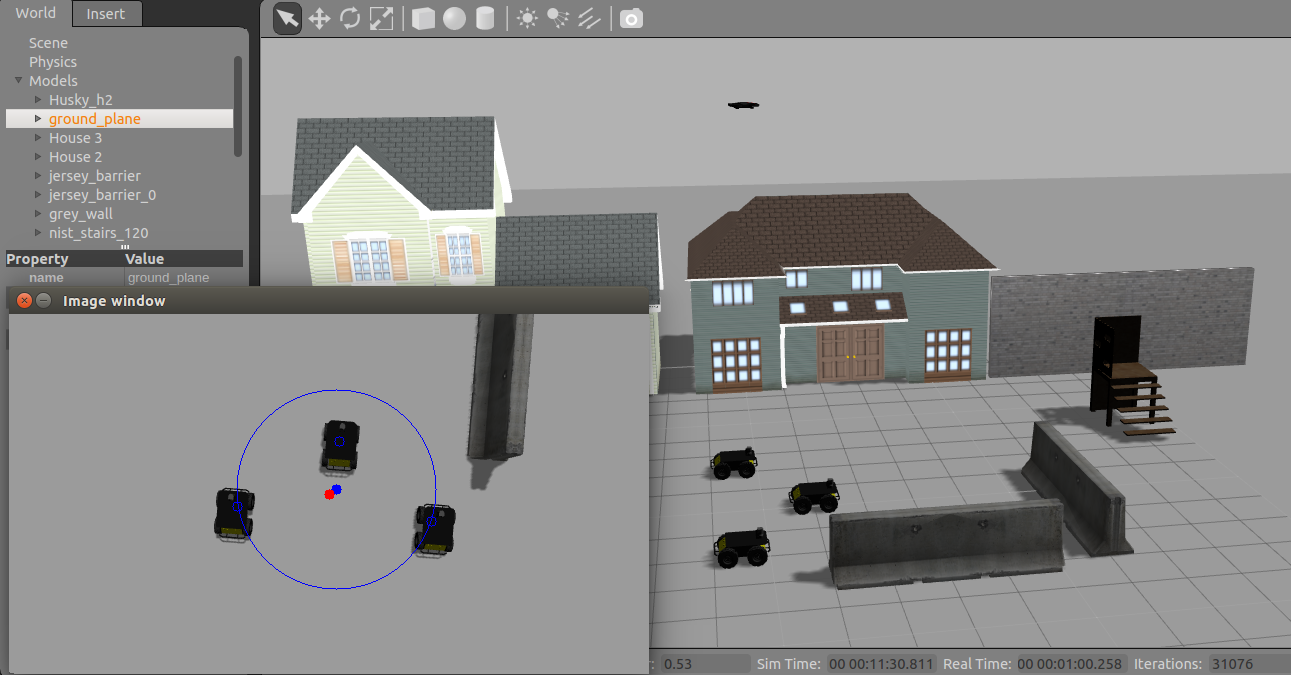}
\caption{Control Environment for UAV and UGVs }
\label{fig:ControlSystem}
\end{figure}

The control environment is shown in Figure~\ref{fig:ControlSystem} in which the red dot is the  centre of the downwards looking camera image from the UAV, and the blue dot and blue circle are the centre of mass and the spread of the UGVs in the UAV's image, respectively. The size of this environment is 10x10 m$^2$

\subsection{Physical Environment}\label{sec:5:1}

Our experimental environment is an in-door UAV testing facility equipped with Vicon. The width of VICON area is approximately 6.6 m and its length is around 5 m. While the UAV and the UGVs travels within the VICON system space, their absolute positions are collected. The origin of coordinates is located at the center of VICON area.

One AR Drone 2.0 and three heterogeneous UGVs (Pioneer P3-AT and P3-DX) are used in our experiments. The agents' posture are directly measured by a Vicon Motion Capture System which broadcasts this information continuously at a high frequency of 100 Hz via UDP protocol. The interaction protocol between the robots and between the ground station (GS) and each robot is achieved using the Robot Operating System (ROS). The wheels on each UGV are equipped with optical encoder sensors to estimate the linear velocity, moving distance and yaw angular rate. An Inertial Measurement Unit (IMU) on the UAV measures angular rates and orientation. Moreover, to guarantee the safety of physical systems, the test space is assigned within the safe area of -3.3 $\div$ 3.3 m width and -2.5 $\div$ 2.5 m length.

\begin{figure}
    \begin{center}
        \includegraphics[width=1\linewidth]{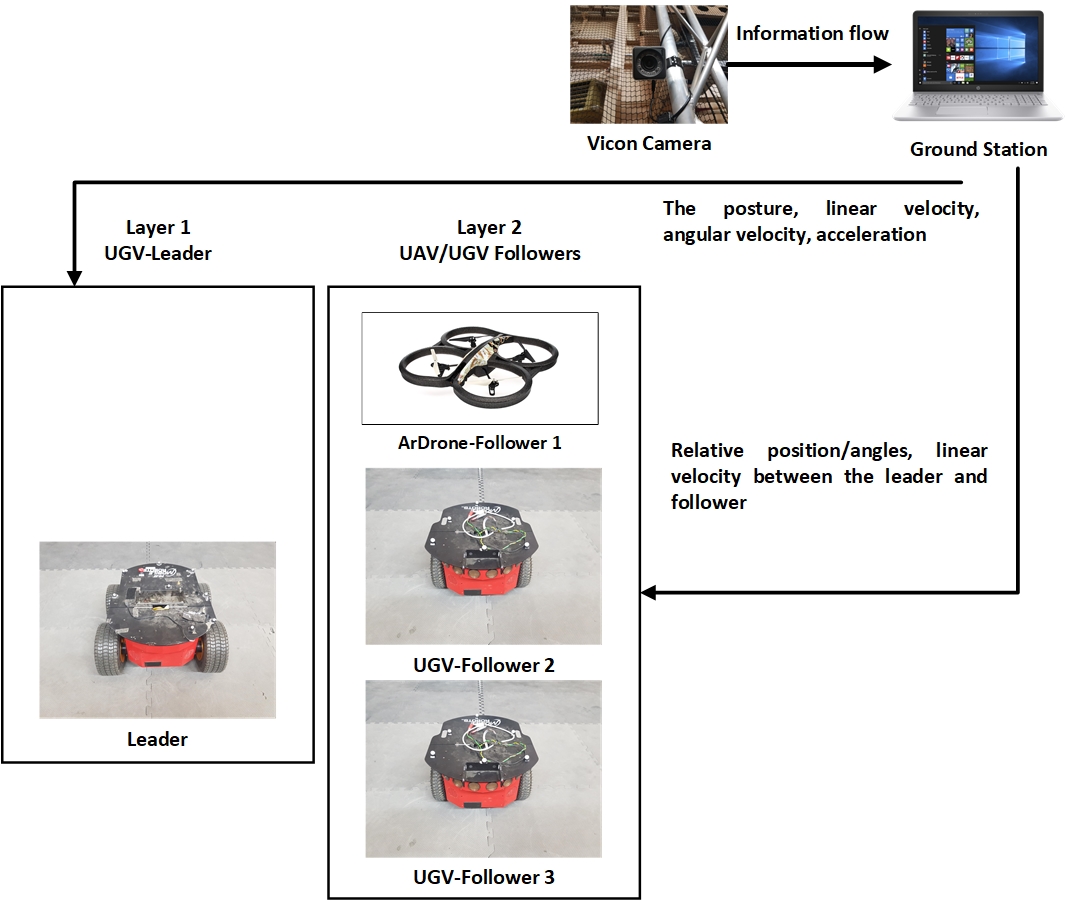}    % The printed column  
        \caption{Overall Architecture Diagram.}  % width is 8.4 cm.
        \label{fig:network_architecture}                                 % Size the figures 
    \end{center}                                 % accordingly.
\end{figure}

The control network architecture is designed to perform cooperative scenarios of the physical UGV-UAV system. The functionality and the task of each block are described in Figure \ref{fig:network_architecture}. Each time step of the system is 10 ms including the time for information exchange, data processing and outputting a command to UAV.

Because of the altitude limitation of the physical space, we just evaluate the control of our UAV agent on the x and y axes. The AI controller calculates the desired velocities on the x and y axes, and sends these directly to the UAV. Additionally, the camera of the AR-Drone was not used in the physical experiment, but instead the central points and radius values were calculated from VICON data directly. 

\section{Experiments}\label{sec:experiment}
In this paper, ABS-DL is evaluted on the UAV-UGVs coordination task. Firstly, the ABS-DL algorithm is evaluated in the simulation environment. Then, our trained DNN model of the ABS-DL algorithm is transferred to the physical environment. During testing the DNN model on the physical environment, the previously described primary safety-net approach is used to avoid dangerous operation of the UAV.

We compare three scenarios: human-combined, where the human performs the complex task, DNN-combined where a DNN is used to learn directly from the human-combined data, and primitive, where the ABS-DL is used to perform the aggregate task by bootstrapping from the sub-tasks. The first scenario is a baseline for human performance on the complex task. The second is a baseline for the AI agent if the data on the complex task could be collected from a human. The third scenario is the proposed ABS-DL where the complex task is bootstrapped from sub-tasks.

For the simulation environment, two setups for autonomous control (2 and 3) are required as described in Table~\ref{tab:experiments}. The first in which a DNN is trained based on human demonstrations of the composite task, called a DNN-combined setup. The second in which a DNN using human demonstrations of sub-tasks or primitive tasks is trained, and then it is tested on the composite task or Combined manoeuvre, called a primitive setup. 

\begin{table}[h]
\caption {Experimental setups} \label{tab:experiments}
 \begin{center}
  \begin{tabular}{cl m{0.6\linewidth}}
    \hline
    ID & Name & Meaning \\\hline 
    1 & Human-combined & Direct human control of the UAV \\
    2 & DNN-combined & Using UAV States Space and demonstration on composite task from human for the Combined manoeuvre \\
    3 & Primitive & Using UAV States Space and primitive demonstrations from human for the Combined manoeuvre \\ \hline
    \hline
  \end{tabular}
 \end{center}
\end{table}

For the physical environment, our trained primitive DNN model is tested under the control of our proposed safety-net approach.  

The UAV's action space consists of four continuous real valued actions representing the pitch, roll, altitude, and yaw, denoted as ($\varphi, \chi, \psi, \omega$), respectively. The state vector of the environment is a 11-D tuple of the continuous variables presented in Table~\ref{tab:StateSpace}.

\begin{table}[h]
\caption {State space}\label{tab:StateSpace}
 \begin{center}
  \begin{tabular}{m{0.15\linewidth} m{0.32\linewidth} m{0.45\linewidth}}
    \hline
    State ID & \centering State name & State description \\ \hline
    \centering 1-2 & $(x^{\mathcal{A}}_t,y^{\mathcal{A}}_t)$ & Centre of UAV from bottom camera\\
    \centering 3-4 & $(x^{\mathcal{G}}_t,y^{\mathcal{G}}_t)$ & Centre of UGV Mass within the UAV image\\
    \centering 5 & $z^{\mathcal{A}}_t$ & UAV's altitude in Gazebo model \\
    \centering 6 & $\nu_t$& Ideal radius of UGVs within image \\
    \centering 7 & $\widehat{\nu}_t$ & Actual radius of UGVs within image \\
    \centering 8-11 & $(\varphi_t, \chi_t, \psi_t, \omega_t)$ & UAV's velocity vector \\
    \hline
  \end{tabular}
 \end{center}
\end{table}

In Table~\ref{tab:StateSpace}, $(x^{\mathcal{A}}_t,y^{\mathcal{A}}_t)$ are received from the UAV's bottom camera, and $z^{\mathcal{A}}_t$, $(\varphi_t, \chi_t, \psi_t, \omega_t)$ obtained from the Gazebo environment. $(x^{\mathcal{G}}_t,y^{\mathcal{G}}_t)$, $\nu_t$, and $\widehat{\nu}_t$ are calculated using the pinhole camera model. $\widehat{\nu}_t$ is the distance from $(x^{\mathcal{G}}_t,y^{\mathcal{G}}_t)$ to the furthest UGV position within the bottom image.

\begin{figure}[h]
\centering
\includegraphics[width=1.0\linewidth]{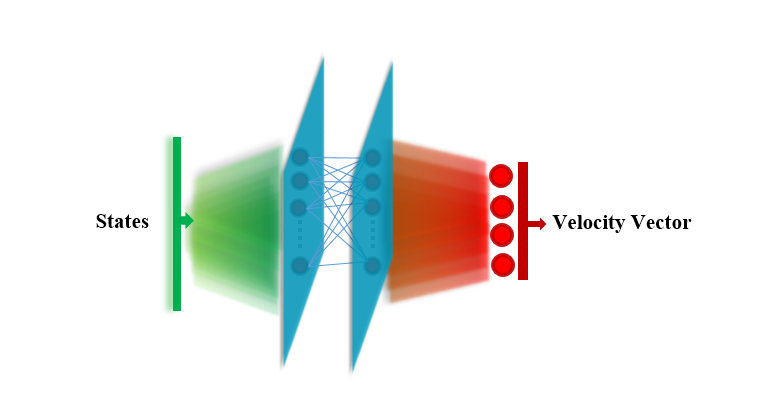}
\caption{Structure of Deep neural network (DNN).}
\label{fig:2}        
\label{fig:DNNarchitecture}
\end{figure}

The DNN architecture used is illustrated in Figure~\ref{fig:DNNarchitecture}. This network consists of an input layer, two fully-connected hidden layers, and a fully-connected output layer, each with 300 nodes. The state space defined in Table~\ref{tab:StateSpace} are used as inputs, while the outputs, as discussed above, are the next continuous action represented by $(\varphi_{t+1}, \chi_{t+1}, \psi_{t+1}, \omega_{t+1})$. The hidden layers use a rectified linear unit (ReLU) as the activation function, whilst the output layer uses tanh. The Adam method~\cite{kingma2014adam} is used for optimization, and mean squared error (MSE) is used for the loss function. Tensorflow and Keras libaries~\cite{chollet2015keras} used to design and train the DNN using a PC with an NVIDIA GeForce GTX 1080 GPU. The weight of layers is initialized at 0.0001.

Demonstrations obtained from the human subject in the simulation environment are collected for 10 episodes in each manoeuvre, except for the Fixed-Altitude manoeuvre in which 5 episodes were performed because the operating path of the UGVs is significantly longer and it is necessary  to balance the labels of the data needed for training DNN. In total, the four data sets have 5296, 4691, 4904, and 5464 instances for
the Fixed-Altitude, Climb, Descend, and Combined manoeuvres, respectively. Meanwhile, each episode runs around 10 minutes for the Fixed-Altitude manoeuvre and 5 minutes for the remaining manoeuvres, and then  each instance is approximate a 0.05sec step These three first data sets are integrated into a primitive data set based on Algorithm~\ref{algh-ABS-DL}. Both of the primitive and combined data set are split into training and validation data sets comprised of 67 percent and 33 percent of the total data. The DNN is trained for 10,000 epochs with a batch size equal to the number of data instances in each setup.

After training in the simulation environment, the DNN for each experiment is tested using testing paths and, in each experiment, the agent is tested ten times on randomly generated cases. 

\section{Results and Discussion}\label{sec:results}
\subsection{In the Simulation Environment}

\begin{figure}[h]
\centering
\includegraphics[width=1.0\linewidth]{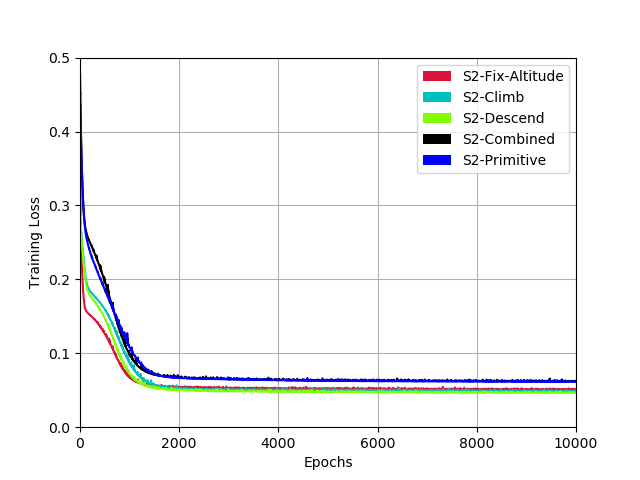}
\caption{Training Loss Measures of Mean Squared Error over Epochs in S2.}\label{fig:S2Loss}
\end{figure}

Figure~\ref{fig:S2Loss} show the value of the MSE in each of the two scenarios for each manoeuvre. In both scenarios, the training is very fast (approximately 40 minutes in real time), with the errors ceasing to decline and becoming stable at approximately 2000 epochs.

To evaluate performances, the two objectives described in  the previous chapter are used. The first (Equation~\ref{eq:objective1}) is the distance between the UAV's Center $(x^{\mathcal{A}},y^{\mathcal{A}})$ and the centre of UGVs' mass $(x^{\mathcal{G}},y^{\mathcal{G}})$, and the second (Equation~\ref{eq:objective2}) is the difference between the actual radius $\widehat{\nu}$ and ideal radius  $\nu$. In Table~\ref{tab:experrors}, the average and standard deviation of these two metrics for the three setups (human combined, combined and primitive) are represented.

\begin{table}[h]
\caption {Averages and Standard Deviations of Errors from Testing in Simulation. The differences are statistically significant at $\alpha=0.05$.} \label{tab:experrors}
  \centering
  \begin{tabular}{m{3cm} cccccc}
    \hline
    \multirow{3}{*}{Experiment ID}  & \multicolumn{2}{c}{} \\
    & Distance Errors & Radius Errors \\      
     & MSE $\mu \pm \sigma$ & MSE $\mu \pm \sigma$ \\
    \hline
Human-combined &   14.2    $\pm$   9   &   12.9    $\pm$   21.9\\
DNN-combined &   12.6    $\pm$   8.1 &   26.2    $\pm$   19.1    \\
Primitive (ABS-DL) &   12.9   $\pm$   8.2    &   10.9 $\pm$ 9.9\\ \hline
  \end{tabular}
\end{table}

The results are interesting. It is evident that the trained DNN using demonstrations of sub-tasks (Primitive) performs much better than the trained DNN using that of the composite task (DNN-combined) regarding the radius error. These results show that our ABS-DL approach can produce policies equivalent or even more effective than the traditional approach in the aerial-ground coordination task.

\begin{figure*}[h]
\centering
 \subcaptionbox{Human-combined}[.3\linewidth][c]{%
   \includegraphics[width=0.3\linewidth]{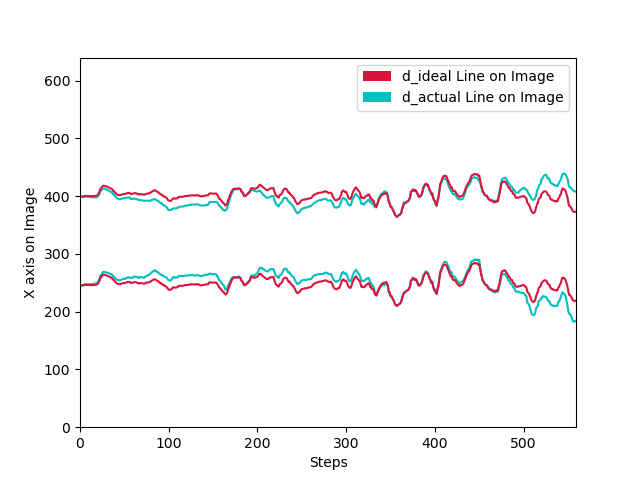}}\quad
  \subcaptionbox{DNN-combined}[.3\linewidth][c]{%
    \includegraphics[width=0.3\linewidth]{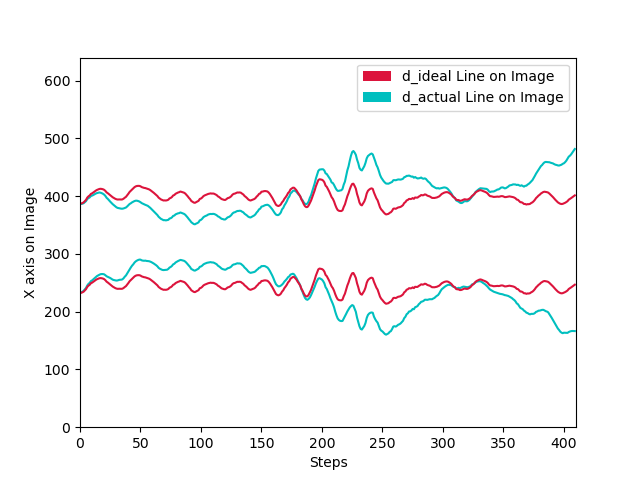}}
  \subcaptionbox{Primitive}[.3\linewidth][c]{%
    \includegraphics[width=0.3\linewidth]{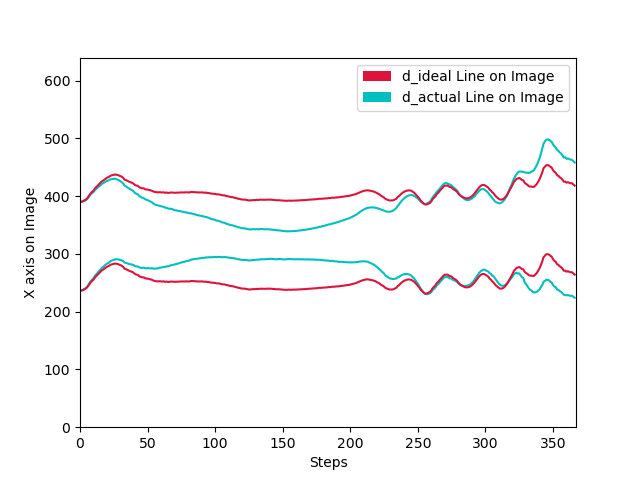}}
\caption{The Ideal and Actual UGVs Circle Trajectories on Horizontal Image in Lateral-Movements-With-Climb-Descend manoeuvre}\label{fig:raxCom}
\end{figure*}

\begin{figure*}[h]
\centering
\subcaptionbox{Human-combined}[.3\linewidth][c]{%
   \includegraphics[width=0.3\linewidth]{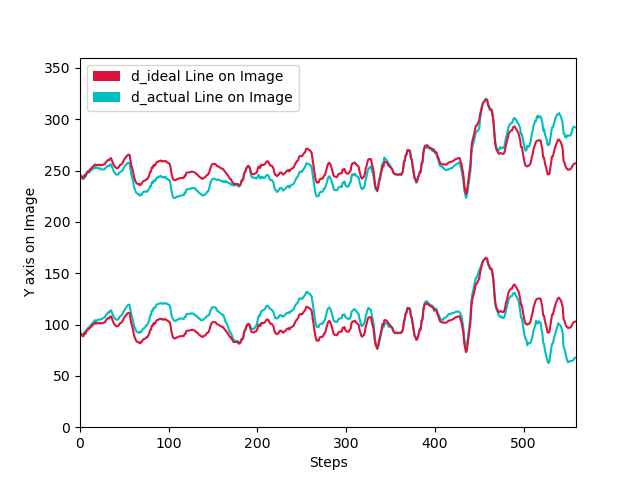}}\quad
  \subcaptionbox{DNN-combined}[.3\linewidth][c]{%
    \includegraphics[width=0.3\linewidth]{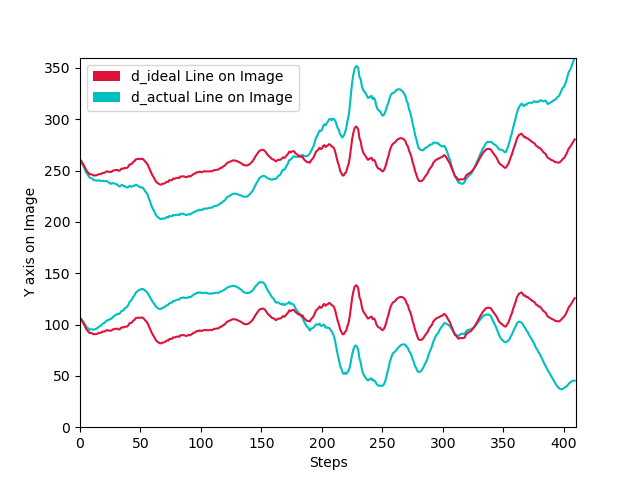}}
  \subcaptionbox{Primitive}[.3\linewidth][c]{%
    \includegraphics[width=0.3\linewidth]{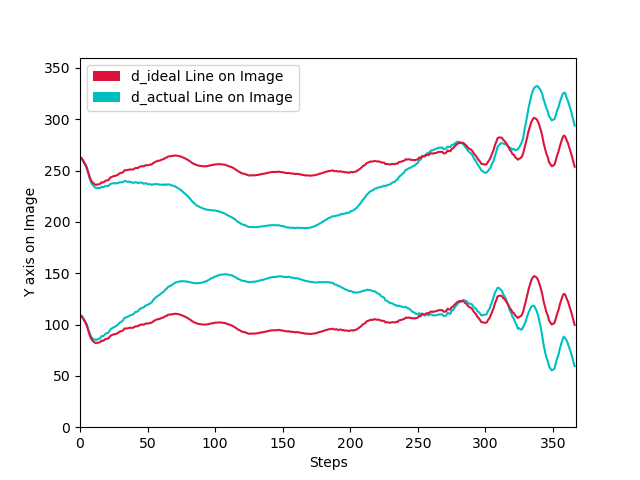}}
\caption{The Ideal and Actual UGVs Circle Trajectories on Vertical Image in Lateral-Movements-With-Climb-Descend manoeuvre}\label{fig:rayCom}
\end{figure*}

It is worth mentioning that despite the variations discussed above, the DNN always retains the UGVs within the range of the camera in all manoeuvres and all test cases. To better understand the phenotypical differences between the human performance and DNN, the visualization of the behaviour of the UAV is shown in Figures~\ref{fig:raxCom}~and~\ref{fig:rayCom} when it is under human control and compare it with the behaviour when it is under combined DNN and ABS-DL control. 

Some general observations can be made based on these figures. Firstly, when tracking UGVs, the DNN seems to do this in a smoother manner, while the human appears to be attempting to track optimally at the cost of generating constant steering of the vehicle. Such a behaviour consumes more energy. Moreover, these figures show that the primitive DNN tracks more smoothly than the combined DNN.   

\subsection{In the Physical Environment}
\begin{figure*}[h]
\centering
\subcaptionbox{Real Positions}[1\linewidth][c]{%
   \includegraphics[width=0.8\textwidth,height=0.25\textheight]{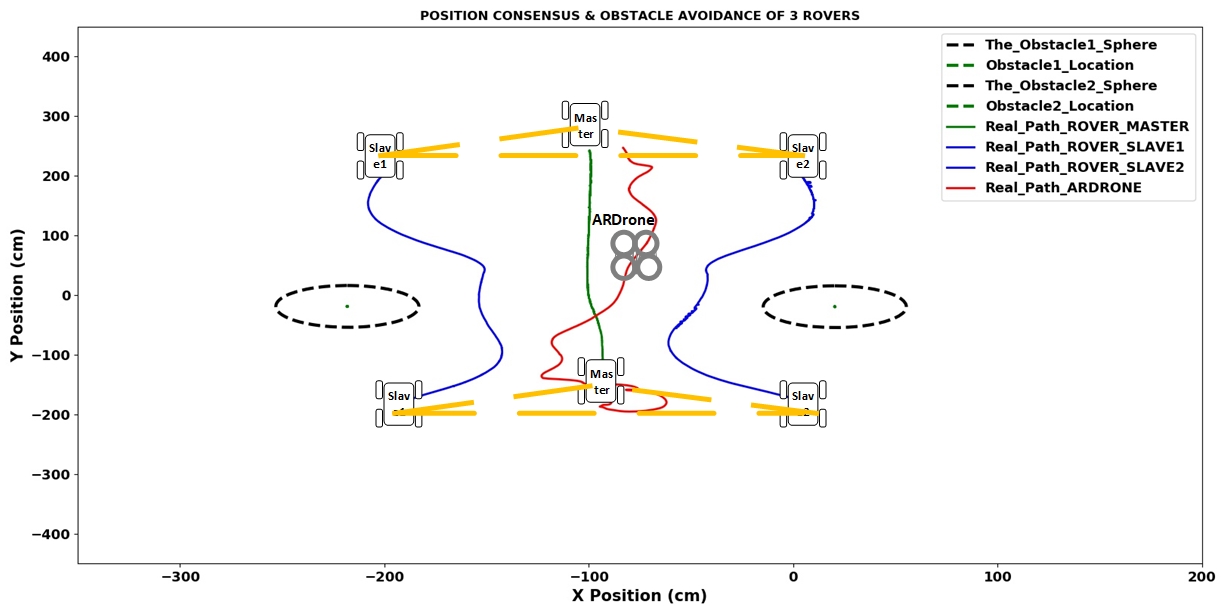}}\quad
   
  \subcaptionbox{Horizontal Image}[.4\textwidth][c]{%
    \includegraphics[width=0.4\textwidth]{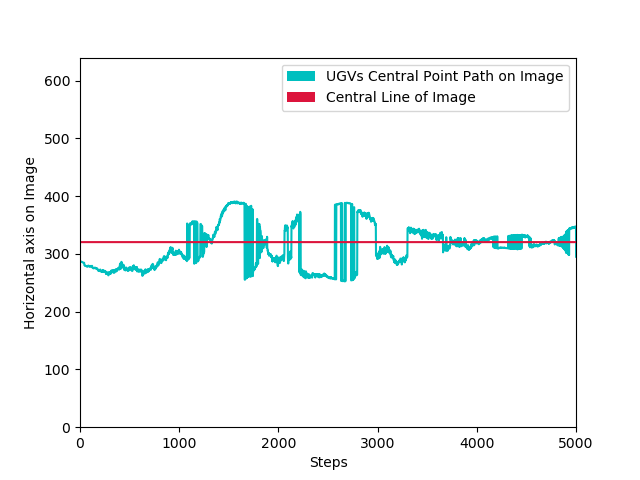}}    
  \subcaptionbox{Vertical Image}[.4\textwidth][c]{%
    \includegraphics[width=0.4\textwidth]{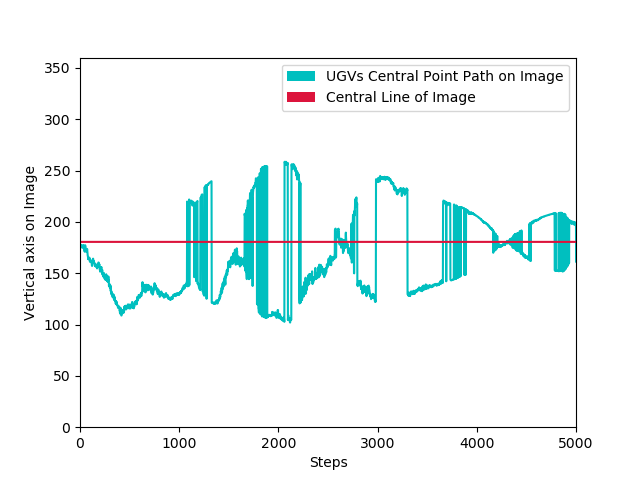}}
    
\caption{UAV and UGVs Trajectories in Physical Environment}\label{fig:Physical_Dis}
\end{figure*}

In this paper, the trained primitive DNN model is tested under the control of our proposed safety-net approach. Figure~\ref{fig:Physical_Dis} shows that the UAV is able to track the UGVs movement when all of the UGVs are within the bottom camera of the UAV. 

It is worth mentioning that the human intervention was minimum compared to flying time. The safety net ensured the UAV to operate within the environmental boundary of the testing facility. However, human intervention were necessary at points of time where the UAV overshots a position.

\section{Conclusion and Future Work} \label{sec:conclusions}

In the simulation environment, results show that the ABS-DL algorithm is able to effectively solve the primary challenge of apprenticeship learning when it produces equivalent or even better policies than that provided by the human operator.

Moreover, when testing in the physical environment, the trained DNN primitive model transferred well and the proposed safety-net approach allowed performance to operate smoothly and for the UAV to track the UGVs movement successfully. These results show that the combination of ABS-DL and the safety-net model in the physical environment is practical and promising.

In the future work, we aim to test different safety-net models for our ABS-DL algorithm on various UAV-UGVs coordination tasks and to completely remove the external intervention.

\section*{Acknowledgement}
This material is based upon work supported by the Air Force Office of Scientific Research under
award number FA2386-17-1-4054.

\bibliographystyle{aaai}
\bibliography{References}

\end{document}